# Review of Methods for Handling Class-Imbalanced in Classification Problems

**Satyendra Singh Rawat [a], Amit Kumar Mishra [b]**

[a] Amity University, Gwalior, India, satyendra.rawat@s.amity.edu

[b] Amity University, Gwalior, India, akmishra1@gwa.amity.edu

**Abstract**

Learning classifiers using skewed or imbalanced datasets can occasionally lead to classification issues; this is a serious issue. In some cases, one class contains the majority of examples while the other, which is frequently the more important class, is nevertheless represented by a smaller proportion of examples. Using this kind of data could make many carefully designed machine-learning systems ineffective. High training fidelity was a term used to describe biases vs. all other instances of the class. The best approach to all possible remedies to this issue is typically to gain from the minority class. The article examines the most widely used methods for addressing the problem of learning with a class imbalance, including data-level, algorithm-level, hybrid, cost-sensitive learning, and deep learning, etc. including their advantages and limitations. The efficiency and performance of the classifier are assessed using a myriad of evaluation metrics.

1. **Introduction**

In the realms of machine learning and data mining, class imbalance learning is a significant problem. In recent years, increasing attention has been paid to the categorization of class-imbalanced data from a variety of fields of study. A balanced sample distribution across classes is generally achieved by traditional classification techniques. However, such a belief led to the majority class performing unfavorably. Since classifiers normally try to reduce the overall classification error, any classifier learned from an imbalanced dataset would exhibit more classification errors in comparison to examples of minority classes (Barua & Murase, n.d.).

We have a better understanding of the nature of imbalanced learning with the arrival of the big data era and the advent of machine learning and data mining, but we are also facing new challenges (Krawczyk, 2016). As in data mining and machine learning communities, finding rare events can be seen as a prediction task. The prediction task suffers from a lack of balanced data because these events are scarce in daily life (Haixiang et al., 2017). Due to the diverse and complex structure of the significantly larger datasets, big data makes it more difficult to lower class disparity. In real-world data, such as fraud detection, spam detection, software defect prediction, etc., these unbalanced datasets are very common (Huda et al., 2018).

Detecting electronic fraud in transactions also poses an extremely challenging problem in class imbalance with overlap. In order to avoid scrutiny, fraudsters have spent a lot of effort in closely cloning a legitimate transaction. It is difficult to distinguish between legitimate and illegal transactions due to the huge amount of data that overlap. For machine learning-based fraud transaction detection methods, overlapping problems have, however, received less attention than problems with class imbalance (Li et al., 2021).

The rationale for the imbalanced data is biased in favor of the majority of class instances owing to high training accuracy. The generation of data from the minority class is consistently regarded as the solution to the issue that has the best chance of success (Dong et al., 2022).

**1.1. Class-Imbalance Problem**

Learning classifiers from skewed or unbalanced datasets is a common problem in classification problems, which is a serious problem. In these situations, the majority of instances belong to one class, while the other class, which typically comprises the more important trait, actually accounts for a significantly lower number of instances. Traditional classifiers generally categorize all of the data into the majority class, which is typically the class with the lowest importance, leaving them obviously unsuited to handle unbalanced learning tasks (Kotsiantis et al., n.d.).

A population with rare diseases, for example, can have medical data with few disease categories. Statistical and machine learning techniques are prone to encounter issues when some classes are glaringly underrepresented. Despite being learned, cases from the rare classes are lost amid the others. The resulting classifiers misclassified unknown rare cases, and descriptive models could have misrepresented the data. If a small class is difficult to identify due to its other characteristics, the learning task becomes significantly more difficult. A small class, for instance, may significantly overlap the other classes. The following depicts a small, difficult class as an interesting class numerous domains exhibit class imbalance, including fraud detection, spam filtering, disease prediction, software defect prediction, ransomware, detection, etc. (Laurikkala, 2001).

This paper discusses the various techniques that are used to handle the class imbalanced data sets in binary classification problems and also provides a comparative study of the most popular methods with their benefits and limitations. The remaining portions of the paper are as follows: In section 2, the literature review contains a few important methods of class-imbalance learning. Existing methods are described in section 3. In section 4, important evaluation metrics are discussed. Finally, the conclusion is given in section 5.

## 2. Review of Literature

The author in this work has discussed current research challenges related to learning from imbalanced data that have roots in contemporary real-world applications and also analyzed different aspects of imbalanced learning, such as mining data streams, clustering, classification, regression, and big data analytics, and given a thorough overview of new challenges in these fields (Krawczyk, 2016). An open-source Python toolbox called imbalanced-learn aims to offer a variety of solutions for the imbalanced dataset issue that frequently arises in pattern recognition and machine learning (LemaîtreLemaître et al., 2017).

Sampling techniques like the synthetic minority oversampling technique (SMOTE) have been applied to unlabeled data to artificially balance the dataset for classifier training. In this paper, a weighted kernel-based SMOTE (WK-SMOTE) that oversamples the feature space of the support vector machine (SVM) classifier is implemented to resolve SMOTE's limitation for nonlinear problems (Mathew et al., 2018). Unfortunately, defective modules typically have a lower presence in software defect datasets than non-defective modules. For imbalanced software defect datasets, the MAHAKIL synthetic oversampling method is introduced, which is based on the chromosomal theory of inheritance (Bennin et al., 2018).

An RK-SVM algorithm based on sample selection was proposed to address the class-imbalance issue in the identification of breast cancer (Cheng et al., 2019). Noise and borderline cases are two important problems brought on by SMOTE's blind oversampling. In (Hussein et al., 2019) proposed the advanced SMOTE, also known as A-

SMOTE, according to the distance between the newly introduced minority class examples and the original minority class samples.

Due to a high-class imbalance, random undersampling using conventional binary classifiers was unable to converge to a sufficient solution for the fraud detection problem. A new method using entropy-based undersampling interlaced with a dynamic stacked ensemble was developed after evaluating the class imbalance problem using a variety of methods (Laveti et al., 2021). Minority class data are transformed into a realistic data distribution when the minority class data are insufficient for GAN to process them effectively on its own (Sharma et al., 2022). The controlled sampling method QDPSKNN used in this study was developed to account for the uneven class distribution of user click data in the classification of fraudsters (Sisodia & Sisodia, 2022).

## 3. Existing Methods

The methods to handle the problem of class imbalance can be divided into four broad categories:

### 3.1. Data-level (or Resampling) Methods

Changes to the training set's distribution are made using data-level techniques, which keep the algorithm's overall structure, including the loss function and optimizer, undisturbed. In order to make popular learning algorithms, data-level methods try to alter the dataset (Kotsiantis et al., n.d.).

Resampling is a method that balances the number of majorities and minority instances in training data. Undersampling techniques and oversampling techniques are the two kinds of resampling methods. Figure 1 depicts the concept of resampling is given below.

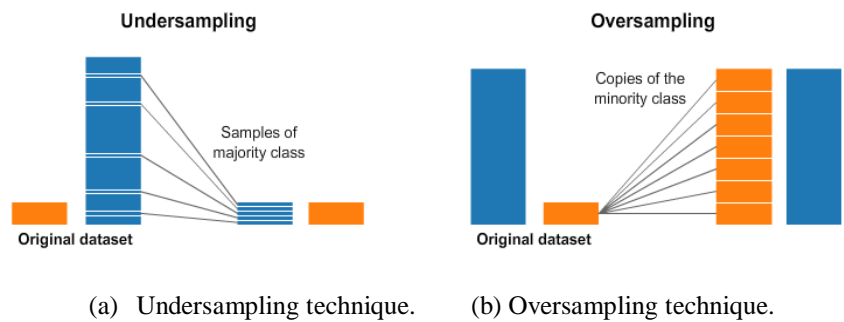

(a) Undersampling technique.    (b) Oversampling technique.

**Figure 1**. **Resampling methods (Agarwal, R. 2020).**

In undersampling methods, by deleting a portion of the majority examples from the training data, an undersampling enables the balance of the majority and minority occurrences. During undersampling, the majority of class samples are removed one at a time until the size of the two classes is nearly equal. As seen in Figure 1 (a).

This review paper provides a comprehensive study of advancements in the classification of unbalanced data. Emerging with several examples of application domains that the class imbalance problem disturbs, this paper discusses the problem's nature. It reviews the most familiar classifier learning algorithms, such as decision trees, backpropagation neural networks, Bayesian networks, nearest neighbors, support vector machines, and associative classification, in order to gain insight into how challenging it addresses these algorithms to learn from imbalanced data (Sun et al., 2009). In Table 1, the few significant undersampling methods are compared along with their advantages and limitations.

| Undersampling Methods | Dataset | Performance metrics | Compare algorithm(s) | Advantages | Limitation |
|---|---|---|---|---|---|
| Relevant Information-based UnderSampling (RIUS) (Hoyos-Osorio et al., 2021) | Glass, Haberma Iris0, Vehicle, Yeast, Pima, Ecoli. | Sensitivity, Specificity, G-mean, AUC | RUS1, UB4, SBAG4 | It chooses the majority class's most pertinent examples. | It is solely appropriate for binary-class tasks. |
| Downsampling for Binary Classification with a Highly Imbalanced Dataset Using Active Learning (Lee & Seo, 2022) | Pima, Haberman, Vehicle, Yeast, Synthetic, Abalone, Poker, Letter, Wine | F-measure, G-mean, AUC, AUC-PR | No sampling, TL, NCL, SMOTE, Random downsampling, Random oversampling | To minimize the impact of imbalanced class labels, it select the samples that were most informative. | It focuses on classification with a binary imbalance. |
| EUStack (Entropy-based undersampling with dynamic stacked ensemble model) (Laveti et al., 2021) | It contains credit card transactions made by European cardholders in September 2013 | Precision, Recall, F1-score, MCC | AdaBoost, Gradient Boost, XGBoost, LDA, Naïve Bayes, Stacked Ensemble | Picks the subset of samples from the dominant class that is most informative. | It can serve as a fraud detection method. |

**Table 1. Undersampling methods.**

As you can see once more in Figure 1(b), an oversampling process makes a similar proportion of synthetic minority samples to original minority samples until the sizes of both classes are almost equal. A few significant oversampling techniques are shown in Table 2 along with their advantages and limitations.

| Oversampling | Dataset | Performance metrics | Comparative algorithm(s) | Advantages | Limitation |
|---|---|---|---|---|---|
| WK-SMOTE (Weighted Kernel-based SMOTE) (Mathew et al., 2018) | Pima, Segment0, iris0, yeast, glass, ecoli. | G-mean | SVM, SMOTE, Borderline, AdaSyn, PI-SMOTE, SVMDC, SMOTEDC, | It balances the class distribution in an SVM classifier. | It is mainly introduced for real-world industrial fault detection problems. |
| MAHAKIL (Bennin et al., 2018) | Ant, arc, camel, ivy, jedit, log4j, pbeans2, redactor, synapse-1.0, | pf measure | SMOTE, Borderline-SMOTE, ADASYN, Random Oversampling | | It does not work in local patches for multi-cluster datasets. |
| GSMOTE-NFM (grouped SMOTE algorithm with noise filtering mechanism) (Cheng et al., 2019) | Pima, Haberman, Wisconsin, glass, new_thyroid, vehicle, ecoli, | G-mean, F-measure | ROS, SMOTE, SL-SMOTE, GG-SMOTE, RNG-SMOTE, | GSMOTE-NFM algorithm generally has better adaptivity and robustness. | Its time complexity is generally higher than some other oversampling algorithms. |
| SMOTEFUN (Synthetic Minority Over-Sampling Technique Based on Furthest Neighbour Algorithm) (Tarawneh et al., 2020) | Pima, Phoneme, Australian,,Bank, Heart, Oil-Spill, Abalone90,Page-block0. | ROC, AUPRC, Wilcoxon Signed –rank test | SMOTE, ADASYN, SWIM using Naïve Bayes and SVM classifiers. | It does not have parameters to tune (such as k in SMOTE). Thus, it is significantly easier to utilize in real-world applications. | It might suffer, especially if one minority class is isolated from both the minority and majority classes and treated as an outlier. |
| SMOTE-tBPSO-SVM (Almomani et al., 2021) | Ransomware dataset | Sensitivity, Specificity, and G-mean | SMOTE, Borderline-SMOTE1, Borderline-SMOTE2, ADASYN, SVM-SMOTE, | An evolutionary-based machine learning approach for ransomware detection. | It does not utilize more data and advanced models to handle big data. |
| Approx- SMOTE (Juez-Gil et al., 2021) | SUSY IR4, SUSY IR6, HIGGS IR4 | AUC, F1-score | No-sampling, SMOTE-BD | It alleviates problems related to imbalanced learning in Big Data scenarios. | It is designed as an algorithm for the Apache Spark framework |

**Table 2. Oversampling methods.**

The problem of class imbalance is alleviated by existing solutions like undersampling and oversampling, but they still have substantial limitations. Undersampling, for instance, results in the loss of samples containing valuable data related to the majority class, and oversampling necessitates a considerable amount of computational time. The combination of these problems makes it difficult to apply the fraud detection model (Laveti et al., 2021).

The benefits and disadvantages of under and over-sampling-based algorithms are unique to them. It is suggested to use a hybrid resampling algorithm that combines oversampling and undersampling if you want results in data processing that are truly accurate. In reducing the proportion of majority samples while raising the number of minority samples, sample imbalance is largely minimized (Xu et al., 2020). The few significant hybrid methods are listed in Table 3 along with their advantages and limitations.

| Hybrid resampling | Dataset | Performance metrics | Compare algorithm | Advantages | Limitation |
| --- | --- | --- | --- | --- | --- |
| RFMSE (Xu et al., 2020) | Spambase, abalone, Contraceptive diabetes, balance, haberman, | Sensitivity, Specificity, F-value, MCC | SMOTE (SM), CCR, GSM, KSM, IHT, RBU, SMOTE-ENN | It is used to handle data imbalance in medical diagnosis. | It still has a very large gap in the medical diagnosis thinking process of doctors. |
| RK-SVM (Random Over Sampling Example, K-means and Support vector machine) | Pima, Transfusion, Iris | Accuracy, sensitivity, specificity, G-mean, AUC, MCC | RK-boosted C5.0, R-SVM, R-boosted C5.0 | It improves performance significantly without increasing algorithm complexity. | In the reality, the data label is very expensive to obtain. |
| SA-CGAN (Single Attribute guided Conditional GAN) (Dong et al., 2022) | Contraceptive, Wine, Dermatology, Yeast | Recall, Precision, Accuracy, F1-score | GAN, CGAN, SMOTE, ADASYN, SVM, K-NN, LR, DT | Avoid unclear, noisy synthetic samples and over-fitting problems. | Some local information on certain data attributes didn't explore. |
| SMOTified-GAN (Sharma et al., 2022) | Connect4, Credit-card, Fraud, Shuttle, Spambase | Precision, Recall, F1-score | Non-oversampled, SMOTE, GAN | Its time complexity is also reasonable for a sequential algorithm | It is an offline pre-processing technique. |
| Hybrid bag-boost model with K-Means SMOTE–ENN (Puri & Gupta, 2022) | Glass, Ecoli, Yeast | AUC, Friedman test, Holm's test | SMOTE, SMOTE-ENN, K-Means-SMOTE | Hybrid bag-boost model for handling noisy class imbalance datasets. | It is only working for binary class noisy imbalanced datasets. |

**Table 3. Hybrid methods.**

### 3.2. Algorithm-level Methods

This study discussed a new method to unbalanced classification it utilizes a single-class classifier technique to accurately capture the properties of the minority class (Kotsiantis et al., n.d.). The RUSBoost algorithm is described by (Seiffert et al., 2010), as a novel hybrid sampling/boosting method for learning from skewed training data and this technique is used in place of SMOTEBoost (Chawla et al., n.d.), another technique that mixes boosting and data sampling. In this work, a new technique for classification noisy label-imbalanced data is proposed, based on the bagging of Xgboost classifiers (Ruisen et al., 2018). Weighted Ensemble with One-Class Classification with Over-sampling and Instance Selection (Czarnowski, 2022) is the name of the proposed method that combines a weighted ensemble classification with a method to tackle the issue of class imbalance (WECOI).

### 3.3. Cost-Sensitive Learning

Cost-Sensitive Learning (CSL), which take into account the different misclassification costs for false negatives and false positives, seems to be another helpful method (López et al., 2012). In (Khan et al., 2015) proposed a cost-sensitive (CoSen) deep neural network which can automatically learn acceptable feature representations for both the majority and minority classes. The results of experiments indicate that the function fitting strategy is more efficient than grid searching in obtaining the optimal cost weights for datasets showing imbalanced gene expression (Lu et al., 2019).

Cost-sensitive Feature Selection Combining the GVM and BALO algorithms, the General Vector Machine (CFGVM) algorithm solves the imbalanced classification problem (Feng et al., 2020). The Correlation-based Oversampling aided Cost Sensitive Ensemble learning (CorrOV-CSEn) is a proposed method that incorporates correlation-based oversampling and the AdaBoost ensemble learning model. While the AdaBoost model includes a misclassification ratio-based cost-function to allow adaptive learning of imbalanced cases, correlation-based oversampling generally includes selecting a suitable oversampling zone and specifying an oversampling rate (Devi et al., 2022).

### 3.4. Deep Learning Methods

Despite research efforts, imbalanced data classification is one of the more major challenges in data mining and machine learning, especially for multimedia data. An extended deep learning approach was offered in (Yan et al., 2016) as a solutions to this problem in order to find skewed multimedia data sets of promising outcomes. More information on the deep learning analysis of a software problem with class imbalance are revealed by this survey (Johnson & Khoshgoftaar, 2019). These studies show a data-level perspective and a temporal window technique to handle the uneven human activity from smart homes and make the learning algorithms more sensitive to the minority class (Hamad et al., 2020). The DNN is a great tool for making complex models to obtain vital information for drug discovery studies (Korkmaz, 2020).

## 4. Evaluation Metrics

The performance of a binary classification problem can be stated to use a confusion matrix, like the one in Table 4. The majority class is marked by a negative label ($y_i = 0$), whereas the minority class is marked by a positive label ($y_i = 1$) [40].

|  |  | Truthful value | |
|---|---|---|---|
|  |  | **Positive (T)** | **Negative (N)** |
| **Estimated Values** | **Positive (T)** | True Positive (TP) | False Positive (FP) |
|  | **Negative (N)** | False negative (FN) | True Negative (TN) |

**Table 4. For a binary classification, a confusion matrix.**

The base metrics for evaluation were False Positives (FP), False Negatives (FN), Precision (P), Recall (R), and F1-Score.

$$\text{Precision} = \text{True Positive}/\text{Total Positives} = TP/(TP + FP) \tag{1}$$

$$\text{Recall/ Sensitivity/Hit} - \text{Rate} = \text{True Positive}/(\text{True Positive} + \text{False Negative}) = TP/(TP + FN) \tag{2}$$

$$F1 - \text{score} = 2 * \text{Precision} * \text{Recall}/(\text{Precision} + \text{Recall}) \tag{3}$$

$$\text{Accuracy} = (\text{True Positive} + \text{True Negative})/\text{Total Values} = TP + TN/(TP + FP + TN + FN) \tag{4}$$

$$G - mean = \sqrt{Sensitivity * Specificity} = \sqrt{TP/(TP + FN) * TN/(TN + FP)} \tag{5}$$

When the classes are unbalanced, the Area under the Precision-Recall Curve (AUC-PR), which is a single statistic that summarizes the precision recall (PR) curve, is another useful metric of the prediction's success. When dealing with heavily skewed data, PR curves are suggested as an alternative to the commonly used receiver operating characteristic (ROC) curve, which could present an overly optimistic picture of the performance for an unbalanced dataset. A binary classification problem can then be used to compare various models using this score, with a score of 1.0 denoting a model with perfect ability. The area under the curve (AUC), which is frequently used in many other articles, is also measured for comparison's purposes (Davis & Goadrich, n.d.).

The prediction only receives a high score if it accurately predicts in each of the four confusion matrix categories (true positives, false negatives, true negatives, and false positives), based on the size of the dataset's positive and negative elements, respectively. A statistical statistic is known as the Matthews correlation coefficient (MCC) (Chicco & Jurman, 2020). The model is perfect whenever the coefficient is +1; when it is 0 or equal to a random hypothesis; when it is -1, the model is totally failed. Contrary to the F1 score, the MCC metric is more reliable (Lee & Seo, 2022).

Table 5 gives the short description of an important evaluation metrics used for classifier's performance analysis.

| Metric | Description |
| --- | --- |
| Precision | It Determines how good the classifier is in detecting the fraudulent cases. |
| Recall | It evaluates the quality of a qualifier. |
| Accuracy | It measures an efficiency of the algorithm. |
| F-measure | It qualifies the quality of a classifier for the occasional classes. |
| G-mean (Geometric mean) | It evaluates the performance of a classifier to create a balance between the minority and majority classes. |
| ROC (Receiver Operating Characteristics) Curve | It is used for evaluating the trade-offs between true positive and false positive error rates in the case of classification algorithms. |
| AUC (Area Under Curve) | It represents the area that exists under a ROC curve. |
| ROC Convex Hull | It is used as a method of identifying potentially optimal classifiers. |

**Table 5. Evaluation metrics.**

5. **Conclusion**

In this work, we assessed a few cutting-edge methods for handling class-imbalance classification problems. Every method has advantages and limitations. On imbalanced datasets, a variety of methods are used, such deep learning, context-sensitive learning, algorithm-level methods, and data-level methods. On the training set, data-level methods such oversampling, undersampling, and hybrids are used. Undersampling algorithms incur information loss, while oversampling algorithms suffer overfitting issues. Despite hybrid algorithms are more effective than resampling methods, but are indeed computationally more expensive and difficult to use. Practical use of algorithmic techniques, such as one-class learning and ensemble learning, are applied at the classifier level (i.e., Bagging and Boosting algorithms). To tackle class-imbalance issues in complex datasets, techniques as deep learning and cost-sensitive learning are also used. To assess the classification accuracy and performance of the classifiers, various evaluation metrics are used.